\documentclass[letterpaper, 10 pt, conference]{ieeeconf}  

\IEEEoverridecommandlockouts            

\usepackage[utf8]{inputenc} 
\usepackage[T1]{fontenc}    
\usepackage{hyperref}       
\usepackage{url}            
\usepackage{booktabs}       
\usepackage{amsfonts}       
\usepackage{nicefrac}       
\usepackage{microtype}      
\usepackage{graphicx}
\usepackage{gensymb}
\usepackage{multirow}
\usepackage{float}
\usepackage[margin=1in]{geometry}

\usepackage[colorinlistoftodos]{todonotes}

\title{\LARGE \bf Evaluating Robustness over High Level Driving Instruction \\ for Autonomous Driving}

\author{Florence Carton$^{1,2}$, David Filliat$^{1}$, Jaonary Rabarisoa$^{2}$ and Quoc Cuong Pham$^{2}$
\thanks{$^{1}$ U2IS, ENSTA Paris, INRIA FLOWERS, 
            Institut Polytechnique de Paris, 
            Palaiseau, France
         {\tt\small florence.carton@ensta.fr, david.filliat@ensta.fr}}%
\thanks{$^{2}$ Université Paris-Saclay,
        CEA, List, F-91120,
        Palaiseau, France
         {\tt\small jaonary.rabarisoa@cea.fr, quoc-cuong.pham@cea.fr}}%
}

\begin{document}

\maketitle

\begin{abstract}

In recent years, we have witnessed increasingly high performance in the field of autonomous end-to-end driving. In particular, more and more research is being done on driving in urban environments, where the car has to follow high level commands to navigate.  However, few evaluations are made on the ability of these agents to react in an unexpected situation. Specifically, no evaluations are conducted on the robustness of driving agents in the event of a bad high-level command. We propose here an evaluation method, namely a benchmark that allows to assess the robustness of an agent, and to appreciate its understanding of the environment through its ability to keep a safe behavior, regardless of the instruction.  

\end{abstract}

\section{Introduction}

Autonomous driving can be achieved using two main paradigms: modular pipeline and end-to-end driving. In the case of the modular pipeline, driving is divided into several sub-modules such as: detection (of road lines, other vehicles, pedestrians, etc...), path planning, and control to compute the actions to be performed (accelerate / brake, steer). This approach allows to train each sub-modules separately and therefore offers a finer control on the agent behavior and a more interpretable training. On the other hand, optimizing sub-modules separately can lead to a sub-optimal policy, since all modules are independent. 
In the case of end-to-end driving, the commands are calculated directly from the inputs, allowing direct optimization, and no intermediate steps. 

A lot of recent approaches rely on end-to-end autonomous driving whether through imitation learning \cite{Bojarski2016b, Codevilla2017, Rhinehart2020, Hawke2019, Toromanoff2018, yang2018end} or reinforcement learning \cite{Perot2017, Toromanoff2019, Kendall2018, Liang2018}. In imitation learning the agent learns by imitating expert's behavior, whereas in reinforcement learning the agent learns by trial and error, based on a reward signal that indicates the correctness of its behavior. 

\begin{figure}[t]
    \centering
    \includegraphics[width=0.49\textwidth]{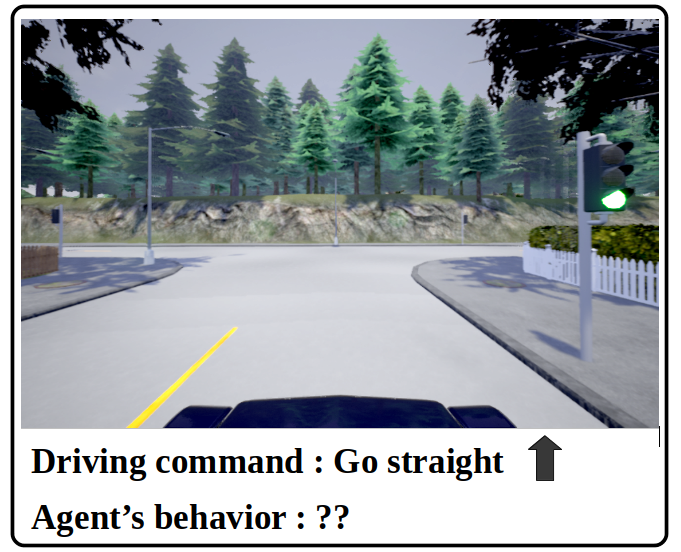}%
    \caption{What to do if the GPS breaks down?}
     \label{fig3}
\end{figure}

While early work on end-to-end driving initially focused on the task of road following, many are now concentrating on driving in urban areas, which involves intersection management. To manage an intersection, the agent must take into account a high-level command (e.g. turn left) provided by a high-level planning system to reach its final goal. 
Nevertheless, the agent must be robust in the sense that the information in the environment must keep priority over the high-level order. This can be for example the case where the gps system indicates to turn, but the street is blocked for road works, or the case where the gps signal is lost due to bad connection and it does not detect that the vehicle is approaching an intersection (Figure~\ref{fig3}). While enforcing such priority is natural in the modular approach, it is not the case for end-to-end approaches and we should therefore evaluate if the controller follows this priority.

In order to highlight the importance of robust decision making of the driving agent, we propose a new way to evaluate their behavior, based on their reaction to mishap. More precisely, we evaluate the stability of an agent over driving instruction, by proposing a benchmark that will evaluate the reaction of the agent if a wrong command is given. We believe these tests are a basis for a necessary evaluation of autonomous driving systems safety. 

We then evaluate several state of the art approaches in this new benchmark, and compare the effect of different training approaches on an agent robustness.

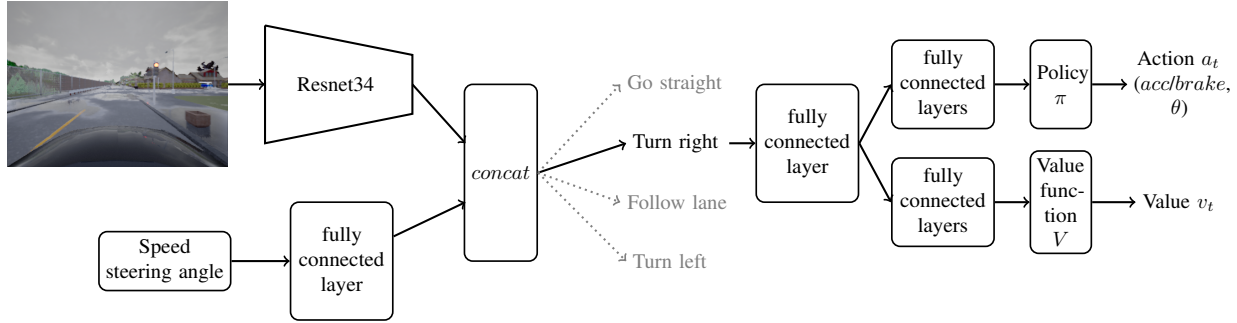
\begin{figure*}
	\centering
	\resizebox{0.99\textwidth}{!}{%
	\begin{tikzpicture}[node distance=1cm, auto,]
	   
	
	\node[inner sep=0pt] (image) at (-4,0) {\includegraphics[height=2.8cm]{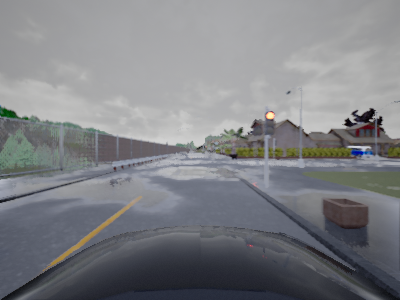}};

	\draw [line width=1pt] [-]  (-1.5,1) -- (1,0.5) -- (1,-0.5) --(-1.5,-1) -- (-1.5,1);
	\node (enc) at (-0.3,0){Resnet34};
	
	\node (measurement) at (-3.2,-3)[draw,thick,minimum width=0.5cm,minimum height=1cm,rounded corners, text width=2.0cm, text centered]{Speed \\ steering angle};
	
	\node (dense) at (-0.2,-3)[draw,thick,minimum width=1cm,minimum height=2cm,rounded corners, text width=1.5cm, text centered]{fully connected layer};
	
	\node (conc) at (2.5,-1.5)[draw,thick,minimum width=0.2cm,minimum height=3cm,rounded corners, text width=1.0cm, text centered]{$concat$};

	\node (b2) at (5.5,-3.0)[text width=1.5cm]{\textcolor{gray}{Turn left}};
	\node (b3) at (5.5,-1.0)[text width=1.5cm]{Turn right};
	\node (b4) at (5.5,0.0)[text width=1.7cm]{\textcolor{gray}{Go straight}};
	\node (b5) at (5.5,-2.0)[text width=1.7cm]{\textcolor{gray}{Follow lane}};

    \node (d2) at (7.7,-1)[draw,thick,minimum width=0.7cm,minimum height=2cm,rounded corners, text width=1.5cm, text centered]{fully connected layer};

	\node (d32) at (10,0.0)[draw,thick,minimum width=0.7cm,minimum height=1.5cm,rounded corners, text width=1.5cm, text centered]{fully connected layers};
    \node (p2) at (12,0.0)[draw,thick,minimum width=0.7cm,minimum height=1.5cm,rounded corners, text width=0.8cm, text centered]{Policy $\pi$};
	\node (a2) at (14,0.0) [minimum width=1cm, text width=1.5cm, text centered]{Action $a_t$ ($acc$/$brake$, $\theta$)};
	
	\node (d42) at (10,-2.0)[draw,thick,minimum width=0.7cm,minimum height=1.5cm,rounded corners, text width=1.5cm, text centered]{fully connected layers};
	\node (vf2) at (12,-2.0)[draw,thick,minimum width=0.7cm,minimum height=1.5cm,rounded corners, text width=0.8cm, text centered]{Value function $V$};
	
	\node (v2) at (14,-2.0) {Value $v_t$};

	\draw [line width=1pt] [->]  (measurement) -- (dense);
	\draw [line width=1pt] [->]  (image.east) -- (-1.5,0);
	\draw [line width=1pt] [->]  (dense) -- (1.9,-2);
	\draw [line width=1pt] [->]  (1,0) -- (1.9,-1);
	\draw [line width=1pt] [->]  (b3.east) -- (d2);

	\draw [line width=1pt, dotted, gray] [->]  (conc.east) -- (b2.west);
	\draw [line width=1pt] [->]  (d2.east) -- (d32.west);
	\draw [line width=1pt] [->]  (d32.east) -- (p2.west);
	\draw [line width=1pt] [->]  (p2.east) -- (a2.west);
	\draw [line width=1pt] [->]  (d2.east) -- (d42.west);
	\draw [line width=1pt] [->]  (d42.east) -- (vf2.west);
	\draw [line width=1pt] [->]  (vf2.east) -- (v2.west);

	\draw [line width=1pt] [->]  (conc.east) -- (b3.west);
	\draw [line width=1pt, dotted, gray] [->]  (conc.east) -- (b4.west);
	\draw [line width=1pt, dotted, gray] [->]  (conc.east) -- (b5.west);

	\end{tikzpicture}

    }
	\caption{Reinforcement learning architecture : model with branch}
	\label{fig:branch}
	\end{figure*}

\section{Related work}

Autonomous driving is a field that has been expanding rapidly in recent years. Since Alvinn \cite{Pomerleau1989} in the 90s, the performance of autonomous cars based on visual inputs has been steadily increasing.  Many approaches use supervised learning \cite{Bojarski2016b, Pomerleau1989, Codevilla2017}, where the agent learns to imitate the behavior of an expert. However we can observe an increasing interest for autonomous driving using reinforcement learning \cite{Kendall2018,Liang2018, Toromanoff2019,Agarwal2019a}.

But whereas a lot of work only handle lane following task \cite{Kendall2018,Bojarski2016b, Wolf2017}, more and more work is being done on the task of driving in urban environments \cite{Codevilla2017, Sauer2018,Toromanoff2019, Liang2018, Chen2019b}. 
With the introduction of Carla simulator \cite{Dosovitskiy2017}, which proposes a number of urban environments, training autonomous driving in the city can be carried out safely in simulation. Different methods are used to indicate to the agent the route to follow. Some use the map and the route to follow (in birdview) \cite{Chen2019c}, other use the waypoints \cite{Agarwal2019} or intermediate goals \cite{Rhinehart2020}, but this requires access to environmental information that is not always available. Finally, one possibility is to give high level commands (straight ahead, turn right, etc...).
To take into account high level control, a branch architecture producing as many output as available high-level commands (Figure~\ref{fig:branch}) is often used \cite{Dosovitskiy2017, Codevilla2017, Toromanoff2019}. It is more effective in forcing the agent to take into account the command \cite{Codevilla2017} than a simple concatenation of the command to the neural network input \cite{Hubschneider2018}. 

Two approaches have recently achieve (almost) perfect score on Carla CoRL Benchmark \cite{Codevilla2017}. The first one with imitation learning named \textit{Learning by Cheating} \cite{Chen2019b} consists in a two stages training. First a privileged agent that has access to the layout of the environment is trained with expert demonstrations. This privileged agent is then used as an expert to train in a supervised way an agent who only has raw image input. 
The second approach called MaRLn \cite{Toromanoff2019}, designed by a team from Valeo, is based on reinforcement learning and affordance. A network is first trained in a supervised way to segment the road and predict information about the environment (distance to next crossing, presence of traffic light, state of this traffic light...). Then the intermediate representations obtained are used as input for a RL agent. This approach reached high performance on the Carla challenge \cite{carla_challenge19, carla_challenge20}. 

As we can see, recent years have witnessed a strong increase in the performance of end-to-end driving, however, as far as we know, no studies have been conducted on the robustness of these approaches. The generalization of driving agents is classically evaluated, by evaluating them in cities that have not been seen, with weather and luminosity conditions different from training ones \cite{Dosovitskiy2017}, or dense dynamic agents \cite{Codevilla2019}. Some have focused on the robustness of driving agents, but usually only deal with the robustness of the detection part \cite{Zhou2018, Michaelis2019}. In \textit{Improved Robustness and Safety for Autonomous Vehicle Control with Adversarial Reinforcement Learning}, Ma et al. \cite{Ma2019} propose a framework to train a robust agent with reinforcement learning, and evaluate it by applying disturbance. The disturbances are applied on acceleration and steering angle. In the same way in ChauffeurNet \cite{Bansal2018}, the agent is trained to be able to handle a trajectory disturbance, and in the Carla Challenge \cite{carla_challenge19,carla_challenge20}, one of the scenarios simulates a loss of control of the vehicle - due to a very wet road. However, no evaluation is made on the reaction of the agents if a wrong command is given.
We propose here a benchmark that would allow an evaluation of the ability of these agents to react in case of unexpected command, and thus measure their ability to understand the environment around them in the sense that they should keep producing a safe control corresponding to the environment situation. We will compare the performance of three of our agents trained using different approaches with two from the state of the art \cite{Chen2019b} and \cite{Toromanoff2019} that achieve high or even perfect scores on Carla CoRL benchmark \cite{Dosovitskiy2017}.

\section{Evaluated agents}

In this section we describe our driving agents. We trained three different agents, all with reinforcement learning. The first one is trained with semantic segmentation as input (as a baseline reaching perfect score on the benchmark), the second with RGB images, and the last one also inputs RGB images but uses auxiliary task to predicts semantic segmentation. Our agents are trained on the version 0.9.6 of Carla simulator \cite{Dosovitskiy2017} and are all evaluated on the original Carla CoRL benchmark with  navigation task. For the purpose of this paper, we do not evaluate the performance on dynamic navigation (i.e. with other dynamic agents - cars and pedestrians) since we only want to evaluate the robustness on wrong commands. 

\subsection{Model and RL training}


We have chosen the PPO algorithm \cite{Schulman2017}, which has proven its effectiveness with input images and continuous actions. PPO belongs to the actor-critic category, and therefore calculates both policy and value function. We use the PPO implementation from \cite{stable_baselines}. 

\subsubsection{Architecture}

\begin{figure}[b]
	\begin{center}
    \resizebox{0.48\textwidth}{!}{%
	\begin{tikzpicture}[node distance=1cm, auto,]
	   
	\node[inner sep=0pt] (image) at (-4.2,0) {\includegraphics[height=3.0cm]{img/input_img.png}};

	\draw [line width=1pt] [-]  (-1.8,1) -- (0.7,0.5) -- (0.7,-0.5) --(-1.8,-1) -- (-1.8,1);
	\node (enc) at (-0.6,0){\Large{Resnet34}};
	
	\draw [line width=1pt] [-]  (3,3) -- (3,4) --(5.5,4.5) -- (5.5,2.5) -- (3,3);
	\node (enc) at (4.2,3.5){\Large{Decoder}};

	 \node[inner sep=0pt] (seg) at (8,3.5) {\includegraphics[height=3.0cm]{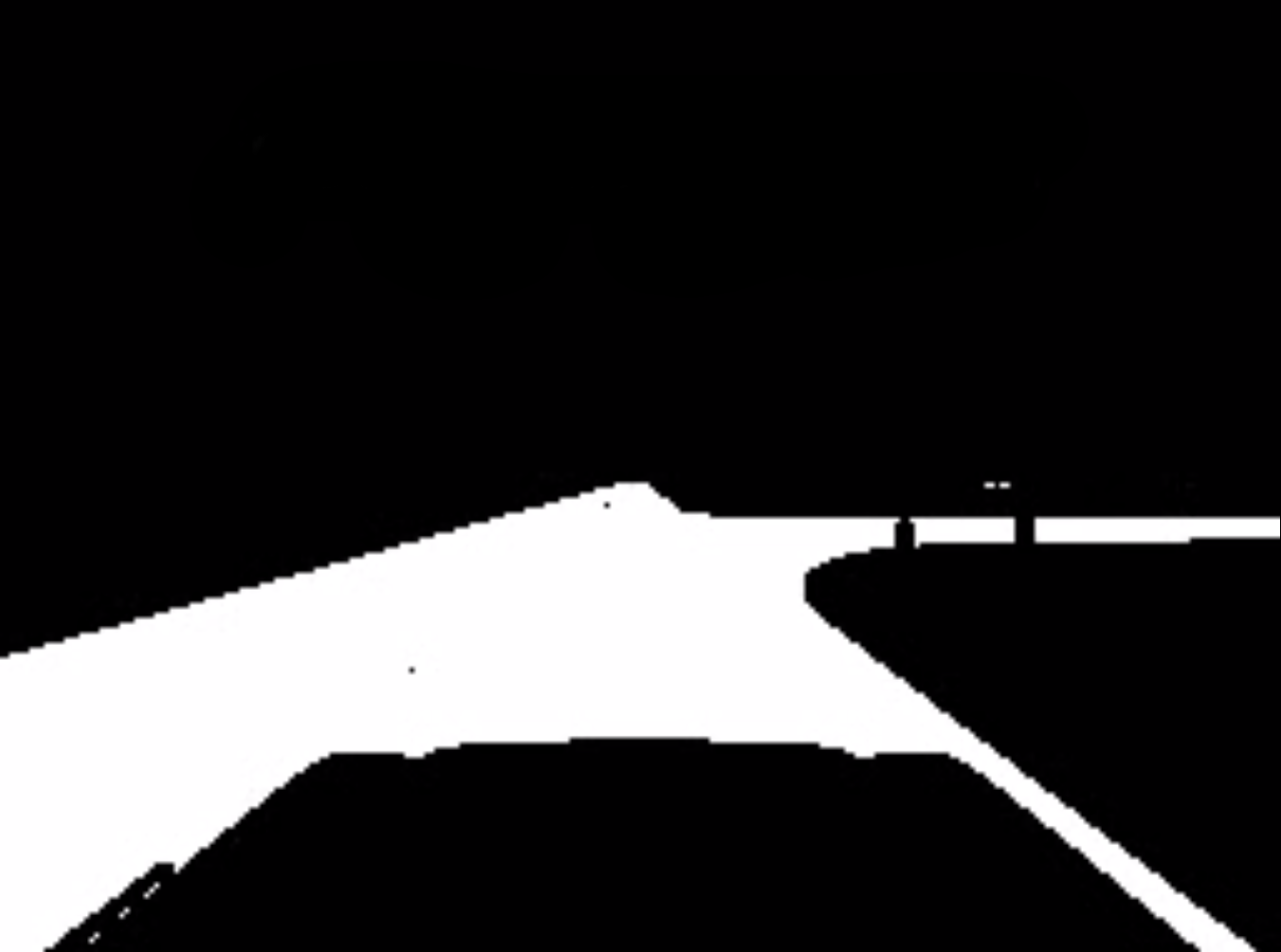}};
	
	\node (measurement) at (-4.0,-3)[draw,thick,minimum width=1cm,minimum height=1cm,rounded corners, text width=3.2cm, text centered]{\Large{speed \\ steering angle}};
	
	\node (dense) at (-0.5,-3)[draw,thick,minimum width=1cm,minimum height=2cm,rounded corners, text width=1cm, text centered]{\Large{FC}};
	
	\node (conc) at (2.2,-1.5)[draw,thick,minimum width=0.2cm,minimum height=3cm,rounded corners, text width=1.4cm, text centered]{\Large{$concat$}};

	\node (b1) at (5.6,-1)[draw,thick,minimum width=0.7cm,minimum height=1.5cm,rounded corners, text width=1cm, text centered]{\Large{FC}};
    
    \node (d31) at (7.5,0)[draw,thick,minimum width=0.7cm,minimum height=1.5cm,rounded corners, text width=1cm, text centered]{\Large{FC}};

	\node (a1) at (9.5,0) [minimum width=1cm, text width=1.5cm, text centered]{\Large{Action $a_t$}};
	
	\node (d41) at (7.5,-2)[draw,thick,minimum width=0.7cm,minimum height=1.5cm,rounded corners, text width=1cm, text centered]{\Large{FC}};

	\node (v1) at (9.5,-2) [minimum width=1cm, text width=1.5cm, text centered] {\Large{Value $v_t$}};

    \node(b2) at (5.1,-2) {};
	\node(b3) at (5.1,-2.5) {};
	\node(b4) at (5.1,-3) {};

	\draw [line width=1pt] [->]  (measurement) -- (dense);
	\draw [line width=1pt] [->]  (image.east) -- (-1.8,0);
	\draw [line width=1pt] [->]  (dense) -- (1.4,-2);
	\draw [line width=1pt] [->]  (0.7,0) -- (1.4,-1);
	\draw [line width=1pt] [->]  (5.5,3.5) -- (seg.west);
	
	\draw [line width=1pt] [->]  (0.7,0) -- (3,3.5);
	
	\draw [line width=1pt] [->]  (conc.east) -- (b1.west);
	
	\draw [line width=1pt] [->]  (conc.east) -- (b1.west) node [midway, above, sloped] (left) {\large{Turn right}};
	
	\draw [line width=1pt] [->]  (b1.east) -- (d31.west);
	\draw [line width=1pt] [->]  (d31.east) -- (a1.west);

	\draw [line width=1pt] [->]  (b1.east) -- (d41.west);
	\draw [line width=1pt] [->]  (d41.east) -- (v1.west);

	\draw [line width=1pt, dotted] [->]  (conc.east) -- (b2.west);
	\draw [line width=1pt, dotted] [->]  (conc.east) -- (b3.west);
	\draw [line width=1pt, dotted] [->]  (conc.east) -- (b4.west);

	\end{tikzpicture}
	}
	\end{center}
	\caption{Reinforcement learning architecture with auxiliary task}
	\label{fig:aux}
	\end{figure}

Among our 3 agents, the general architecture remains the same (Figure~\ref{fig:branch}). The input (RGB image or semantic segmentation) is passed through an encoder (Resnet34 in the case of an image, and the smaller NatureCNN from \cite{Mnih2015} for the semantic segmentation). RGB image inputs are subject to standard data augmentation (partial erasing,  Gaussian blur, variation of hue, saturation, brightness and saturation). The encoder is trained from scratch for the agents with segmentation as input and the one using auxiliary task, and pretrained with semantic segmentation task for the third one.  Current speed and steering angle, which are passed through fully connected layers, are also used as input. The features are then concatenated and the output of the network is composed of the action and the value. To take into account the high-level instruction, a model with branches is used: the action and value outputs are duplicated for each high-level command, and the output of the corresponding branch is used depending on the command issued to the agent.  For the agent using the auxiliary task, a decoder outputting semantic segmentation is added to the model (see Figure~\ref{fig:aux}). The decoder is inspired from Linknet \cite{Chaurasia2018}, which is a segmentation network based on Resnet34. The auxiliary task is trained at the same time as driving task, 3 blocks of the Resnet34 encoder are common to segmentation and driving, the last block is duplicated and task-specific. Training is carried out with a homoscedastic combination of losses (supervised cross entropy loss for segmentation, and reinforcement loss for driving) from \cite{Kendall2017}. Equation \ref{eq:homo} shows the loss function with $l_{rl}$ and $l_{aux}$ respectively the RL loss and auxiliary loss, and $\sigma_{rl}$ and $\sigma_{aux}$ two trainable parameters.
\begin{equation}
    l = e^{-\sigma_{rl}}\times l_{rl} + \sigma_{rl} + e^{-\sigma_{aux}}\times l_{aux} + \sigma_{aux} .
\label{eq:homo}
\end{equation}

\begin{table*}
 \centering
\begin{tabular}{|c|c|c|c|c|}
\hline
Training		& Training 	& New weather		& New town	& New Town 	\\
            & conditions &                  &           & \& weather \\
\hline       
LBC \cite{Chen2019b}&   \textbf{100\%}   &  \textbf{100\%}  &  98\% &   \textbf{100\%} \\
\hline
MaRLn \cite{Toromanoff2019}&   \textbf{100\%}   &  \textbf{100\%}  &  \textbf{100\%} & 98\%   \\
\hline
Segmentation input 	& \textbf{100\%}& \textbf{100\%}&  \textbf{100\%}	& \textbf{100\%} \\
\hline
RGB image input  	& 	 	82\%	& 	98\%	&  49\%	    & 	40\%	\\
\hline
Auxiliary task  	& 	  90\%	& 	92\%	&  78\%	    & 	68\%	\\
\hline
\end{tabular}
\caption{Results of our agents and SOTA agent on CoRL benchmark with navigation task \label{tab:res} }
\end{table*}

\subsubsection{Reward function}

As the reward is the only information the agent obtains from the environment, it is a very important component in reinforcement learning algorithms. 
The reward function used is a weighted sum of three terms : 

\begin{equation}
   r = w_s \times r_{speed} + w_{cte} \times r_{cte} + w_a \times r_{angle}
\end{equation}

with 
\begin{itemize}
    \item $r_{speed} = target\_speed - |target\_speed- speed|$ : Computes the gap between target speed and current speed. Target speed is 35~km/h in general, and 15~km/h at intersections.
        
    \item $r_{cte} = (1 - d) $ with $d$ the cross-track error, i.e. the lateral distance (in meters) of our agent to road center (Figure~\ref{fig2})
    \item $ r_{angle} = (15 - |\alpha|)$ with $\alpha$ the angle (in degree) between the agent's direction and the road direction (Figure~\ref{fig2}). Constants are added in $r_{cte}$ and $r_{angle}$ so that the corresponding reward is positive if the car is roughly aligned with the road (ie $\alpha < 15 \degree $) and not too far from road center ($d < 1 m$)  
    
    \item $w_s$, $w_{cte}$ and $w_a$ their associated weights, respectively $1.0$, $10$ and $0.1$. The weights for every component were determined experimentally.
    
    \begin{figure}[H]
        \centering
        \includegraphics[width=0.27\textwidth]{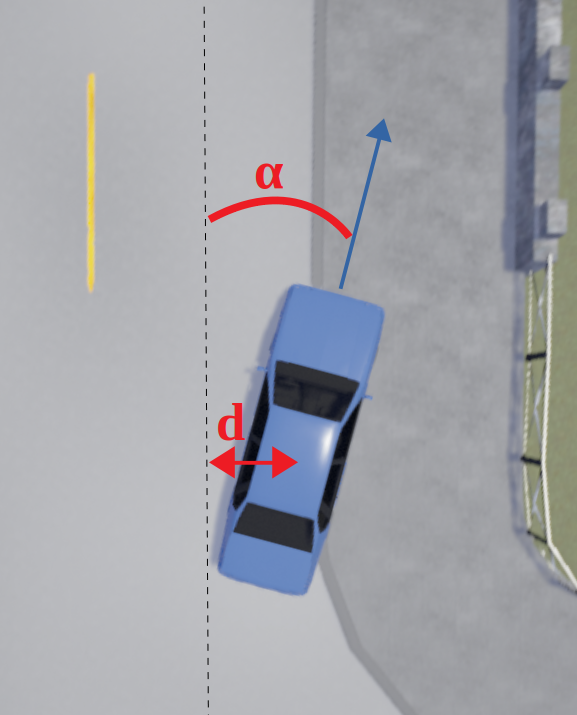}%
        \caption{Illustration of cross-track error $d$ and angle $\alpha$ used to compute the reward}
        \label{fig2}
        \end{figure}

\end{itemize}

\subsection{Results on CoRL Benchmark}

CoRL Benchmark \cite{Dosovitskiy2016} is a goal directed urban navigation benchmark. Four different tasks can be evaluated (from going straight to full navigation with dynamic obstacles), and we will focus on navigation task (navigation without other cars or pedestrians). CoRL benchmark evaluates the generalization capabilities of the agent, by testing training conditions as well as unseen town and new weather conditions. The benchmark outputs the percentage of successful episodes, i.e. when the driving agent has managed to reach destination. For every episode, the trajectory is computed by the simulator and high level commands are sent to the agent with the different information (road picture, current speed, etc...). In the CoRL benchmark, given commands are always correct.

We present in Table~\ref{tab:res} the performance of our agents, and compare them to the state of the art algorithms \textit{Learning by Cheating} (LBC) \cite{Chen2019b} and MaRLn \cite{Toromanoff2019} on the navigation task. We used the open-source implementations of the benchmark released by \textit{Learning by Cheating}\footnote{\url{https://github.com/dotchen/LearningByCheating}} and \textit{MaRLn}\footnote{\url{https://github.com/valeoai/LearningByCheating}}.
For the latter, in order to make a fair comparison, we used the agent who was trained only on training conditions for the CoRL benchmark.  The score is the percentage of success of requested trajectories. Evaluation is made with the version 0.9.6 of Carla. 

We find that among our agents, the one which competes with the state of the art is the one which uses semantic segmentation as input. When using RGB images as input, the performances fall especially in the city that was not seen during the training, although the use of the auxiliary task allows for better generalization. The lower results of our agents with image input in training conditions compared to new weathers are due to a harder weather (hard rain) among training weathers compared to the test weathers. However, the results of our agents with the input image remain high in the training city, with more than 90\% successful episodes on training town for our agent with auxiliary task.

\section{Wrong Command Benchmark}

\begin{table*}[ht]
\resizebox{\textwidth}{!}{
\begin{tabular}{|c|c|c|c|c|c|}
  \hline
 &Type of crossing &  \begin{minipage}{.15\textwidth}
      \includegraphics[width=\linewidth]{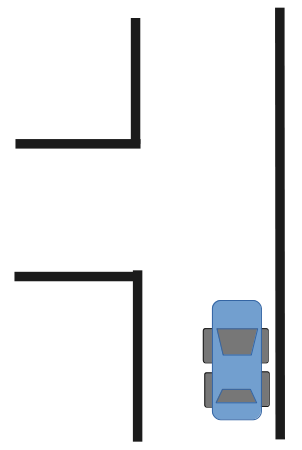}
    \end{minipage}   
    &  \begin{minipage}{.15\textwidth}
      \includegraphics[width=\linewidth]{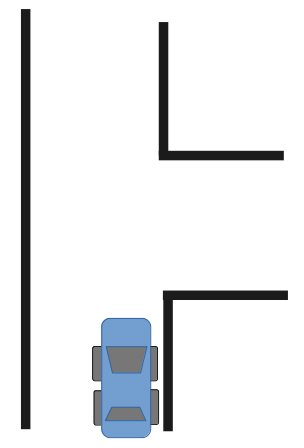}
    \end{minipage} 
    &   \begin{minipage}{.25\textwidth}
      \includegraphics[width=\linewidth]{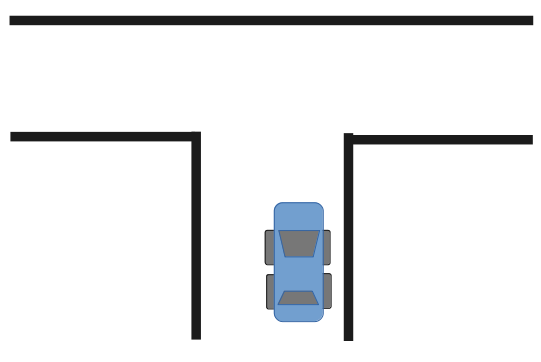}
    \end{minipage} & \\
  \hline
  & wrong command & \begin{minipage}{.05\textwidth}
      \includegraphics[width=\linewidth]{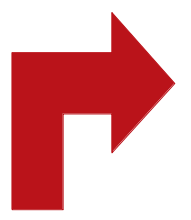}
    \end{minipage}    & \begin{minipage}{.05\textwidth}
      \includegraphics[width=\linewidth]{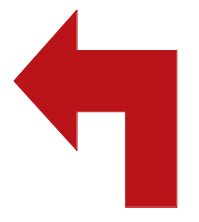}
    \end{minipage}   & \begin{minipage}{.035\textwidth}
      \includegraphics[width=\linewidth]{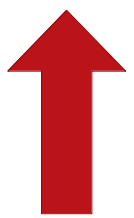}
    \end{minipage}& Total unsafe  \\
  \hline
  \hline
  \multirow{4}{*}{LBC \cite{Chen2019b}} & Unsafe behavior &  \multirow{2}{*}{0} & \multirow{2}{*}{5} & \multirow{2}{*}{10} & \multirow{4}{*}{23 / 60} \\
  & with wrong command &  &   &  & \\
 \cline{2-5}
  & Unsafe behavior & \multirow{2}{*}{0} &\multirow{2}{*}{0} & \multirow{2}{*}{8} &  \\
  & with 'follow lane' command &  &   &  & \\
   \hline
   \hline
  \multirow{4}{*}{MaRLn \cite{Toromanoff2019} } & Unsafe behavior &  \multirow{2}{*}{2} & \multirow{2}{*}{\textbf{1}} & \multirow{2}{*}{10}  & \multirow{4}{*}{13 / 60} \\
  & with wrong command &  &   &   & \\
 \cline{2-5}
  & Unsafe behavior & \multirow{2}{*}{0} &\multirow{2}{*}{\textbf{0}} & \multirow{2}{*}{\textbf{0}} & \\
  & with 'follow lane' command &  &   &  & \\
   \hline
   \hline
  \multirow{4}{*}{RL seg input} & Unsafe behavior &\multirow{2}{*}{0} & \multirow{2}{*}{10} & \multirow{2}{*}{9} & \multirow{4}{*}{35 / 60} \\
  & with wrong command &  &  &  & \\
  \cline{2-5}
  & Unsafe behavior & \multirow{2}{*}{0} & \multirow{2}{*}{6} & \multirow{2}{*}{10} & \\
  & with 'follow lane' command &  &  &  & \\
  \hline
     \hline
  \multirow{4}{*}{RL img input} & Unsafe behavior & \multirow{2}{*}{0} & \multirow{2}{*}{\textbf{1}} & \multirow{2}{*}{7} & \multirow{4}{*}{16 / 60} \\
  & with wrong command &   &  &   & \\
  \cline{2-5}
  &  Unsafe behavior & \multirow{2}{*}{0} & \multirow{2}{*}{4} & \multirow{2}{*}{4} & \\
  &  with 'follow lane' command &  &  & &  \\
  \hline
  \hline
   & Unsafe behavior & \multirow{2}{*}{0} & \multirow{2}{*}{2} & \multirow{2}{*}{\textbf{0}} & \multirow{4}{*}{\textbf{3 / 60}}\\
  RL with & with wrong command &  &  &  &  \\
  \cline{2-5}
   auxilliary task & Unsafe behavior & \multirow{2}{*}{1} & \multirow{2}{*}{\textbf{0}} &  \multirow{2}{*}{\textbf{0}} & \\
  & with 'follow lane' command &  &  &  & \\
  \hline
\end{tabular}
}
\caption{Results of our agents and two state of the art agents on our wrong-command benchmark, showing the number of unsafe behavior on 10 trials.}
\label{tab:benchmark}
\end{table*}

Usually in driving benchmark, tests are made by changing the visual input : different towns and weathers that were not seen during training. This allows to check the generalization capability of the agent. However in autonomous driving, there are not only visual inputs. 
In this section, we want to measure the stability of driving agents over given instructions. 

\subsection{Benchmark Description}

\begin{figure}[t]
    \centering
    \includegraphics[width=0.4\textwidth]{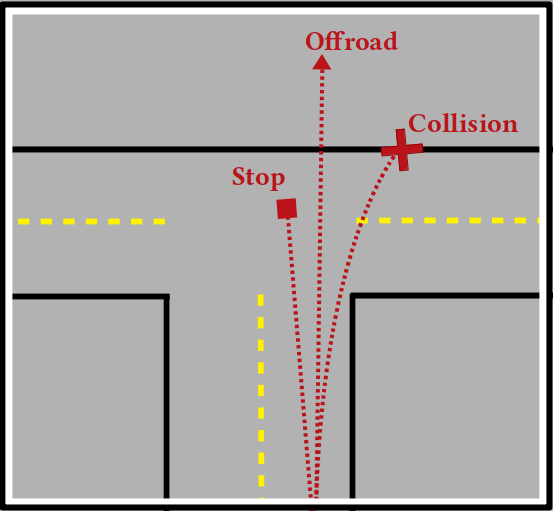}%
    \caption{Examples of unsafe behaviors.}
     \label{xing}
\end{figure}

Our benchmark is inspired by the CoRL benchmark from Carla \cite{Dosovitskiy2017}. It consists in giving an incorrect command at an intersection, and evaluating the reaction of the driving agent. Since we only work in Town01, there are 4 possible commands : follow lane (basic command when there is no crossing), go straight, turn left and turn right. We want to evaluate the reaction of the driving agent if an incorrect command is given at an intersection. 
We can distinguish two cases : either the agent is told to go at a direction that does not exist, or the basis command (follow lane) is still given even if the agent reaches a crossing. We will measure the number of unsafe behavior (i.e. collision, offroad, or stop at the middle of the crossing) in these cases (see Figure~\ref{xing} for examples of unsafe behaviors). This will help to measure if the agent has achieved a certain understanding of the environment, or if it blindly follows instructions. 
For each crossing type, we took 10 of them in the training town, with the ClearNoon time/weather preset, a simple and already seen during training weather condition, and run small episodes starting just before the crossing with erroneous command.

In the event that a wrong command is given at a junction, one can expect two behaviors from a human driver: either it stops before the junction or it randomly takes one of the proposed routes. 
Out of the 10 test episodes, we have therefore counted as unsafe behavior the cases where the agent stops in the middle of the crossing, and the cases where the agent collides or goes offroad. For every crossing type we counted the unsafe behavior 
for both wrong command (that is not available at the given crossing), and basis - follow lane - command. The results are given in table \ref{tab:benchmark} for state of the art algorithm \cite{Chen2019b} and \cite{Toromanoff2019} as well as for our three agents: with the segmentation and with the image in input, and  the last one with the segmentation as auxiliary task. Since the results measure unsafe behavior here, the smaller the more robust. The companion video illustrates some of the agents behaviors\footnote{\url{https://github.com/CEA-LIST/AD-RobustnessEval/blob/master/video/Demo.mp4}}. The code for the benchmark as well as the weights of our RL agent with auxiliary task can be found open-source\footnote{\url{https://github.com/CEA-LIST/AD-RobustnessEval/}}.

\subsection{Qualitative Analysis}

We first note that the results are very disparate according to the type of crossing. For the first crossing type, where the 'turn right' command is not possible, all the agents drive correctly (with sometimes bites into the sidewalk, but overall safe behaviors). This is probably due to the fact that the sidewalk is always present to the right of the agent, and that this helps the agent to drive correctly. 

For the second crossing, where the 'turn left' command is not possible, the agents have different behaviors. 
The LBC \cite{Chen2019b}  agent deviates slightly if it is still asked to turn left, and then in half of the cases, stops abruptly in the middle of the crossing, which we considered to be dangerous and therefore unsafe behavior. The MaRLn \cite{Toromanoff2019} agent keeps its course, and both \cite{Toromanoff2019} and \cite{Chen2019b} perform well when the command remains to follow the road. 
Among our agents, the behavior varies greatly. The agent using input segmentation has a very high collision rate, because it turns directly left if this command is given. The two agents using the input image react better, even though the auxiliary task seems to help to behave correctly. 

Finally the third type of intersection, where it is not possible to go straight, appears to be the more difficult for all agents. The LBC agent has a very bad error management : in the vast majority of cases, it keeps going straight, even if there is a wall, and usually hit any obstacle in front of the car, and so does our agent that uses segmentation. The MaRLn agent manages to turn when the command follow lane is still given - by sometimes biting into the other line, but keeps going straight if the command straight is given. However, unlike LBC, it rarely collides, preferring to stop in the middle of the crossing. On the other hand, our agent which has learned with an auxiliary task does not make any mistakes. It turns arbitrarily to one side or the other, even if it is asked to go straight, which shows that it relies much more on the analysis of the environment around.  

\subsection{Discussion}

We present the performance of 5 driving agents: one trained with imitation learning (LBC \cite{Chen2019b}), and the four others with reinforcement learning. We are not surprised by the rather low performance in terms of robustness of LBC. This is one of the drawbacks of supervised learning, since the cases of error are not present in the learning dataset, the agent never learns how to react to unexpected situations. 

One might have thought that reinforcement learning, because of its trial-and-error nature, would alone have increased the robustness of learning. However, the disparity in the results of agents trained through reinforcement learning shows that this is not enough. 
We can compare the performance of our two agents with RGB image as input. Although using an encoder pretrained on the segmentation task, the simple agent (i.e. without auxiliary task) is less robust than the one using segmentation as an auxiliary task. The use of semantic segmentation as auxiliary task allows the agent to make the link between the semantic parts of the image and the rewards, and thus to have a better understanding of the environment. It is able to analyze its surroundings, and to make a decision in a situation it has never seen before by giving priority to the visual information over the high-level command. 

MaRLn agent \cite{Toromanoff2019}  is quite close to ours using auxiliary task, with the following differences : the encoder is pretrained to calculate segmentation as well as other affordances (distance to traffic lights, position in the road...), and frozen during RL driving training. Perception is then decoupled from driving. Our agent on the other hand is trained end-to-end with both segmentation and driving. We believe that decoupling perception and control may reduce the robustness of the agent.

We have also analyzed the different experiments of our agent using only segmentation, and a possible explanation for the poor performance in terms of robustness is that it is too confident on the road marking perception. Indeed, the segmentation in Carla being perfect, this agent has never needed to use anything other than the road lines to learn to drive. In normal situations, when it reaches a crossroads, our agent starts to turn in the prescribed direction, and adjusts its trajectory with the lines of the road it joins (road markings are not present in the intersection). However, in the case of a wrong command, it starts to turn and cannot correct itself because there are no more lines - because the target road doesn't exist.

On the other hand, the agents which learned with an RGB image as input were subject to data augmentation. In particular, road lines are not always visible in the image (very rainy weather or Gaussian blur), so the agents learn to use other elements - the sidewalk for example. These agents therefore learn with information redundancy, which is not the case if only perfect segmentation is used. One lead for future work to validate this hypothesis would be to train an agent with segmentation input, by regularly and randomly removing one or more masks. The agent would then learn to use the different elements present in the observation in a robust way. 

\section{Conclusion}

In this paper we propose a first study to evaluate the robustness to wrong command of a vision based driving agent. With the development of end-to-end conditional autonomous vehicles, we believe this issue is of crucial interest and has not be dealt with so far. We found in this evaluation that an agent appears to be more robust when perception and control are not decoupled during training. We also observed a better robustness when the agent has a high-level understanding of its environment with, for example, the ability to segment the observation, but also when it learns to use the redundancy of the elements present in the observation.  

As future work, We can imagine others tests to evaluate the robustness: changing the visual input by placing the camera at a different position, changing steering wheel hardness, etc... Since all the situations cannot be seen during training, we need to have agents that are capable of adaption and understanding of the environment, in the sense that they must extract enough information to avoid dangerous situations. 


\medskip



\bibliographystyle{IEEEtran}
\bibliography{IEEEabrv,Bib/library.bib, Bib/website.bib}

\end{document}